\documentclass[10pt,twocolumn,letterpaper]{article}

\usepackage[pagenumbers]{cvpr} 

\usepackage[dvipsnames]{xcolor}

\definecolor{cvprblue}{rgb}{0.21,0.49,0.74}
\usepackage[pagebackref,breaklinks,colorlinks,citecolor=cvprblue]{hyperref}

\usepackage{subfiles} %
\usepackage{makecell} %
\usepackage{multirow} %
\usepackage{subcaption}
\usepackage{color,xcolor}
\usepackage{graphicx}
\usepackage{booktabs}
\usepackage{makecell}
\usepackage{colortbl}
\usepackage{pifont}%
\usepackage{algorithm} %
\usepackage{algorithmic} %
\usepackage[switch]{lineno}
\usepackage{amssymb} %

\usepackage[misc]{ifsym}
\usepackage{orcidlink}

\usepackage[accsupp]{axessibility}

\definecolor{shapecolor}{rgb}{0.0,0.5,0.0}
\newcommand{\name}{SAVSSM}%

\definecolor{arylideyellow}{rgb}{0.91, 0.84, 0.42}

\definecolor{cvprblue}{rgb}{0.21,0.49,0.74}
\usepackage[pagebackref,breaklinks,colorlinks,citecolor=cvprblue]{hyperref}

\newcommand{\arch}[1]{\textsc{#1}}

\title{SaMam: Style-aware State Space Model for Arbitrary Image Style Transfer}

\author{Hongda Liu\textsuperscript{1}, 
Longguang Wang\textsuperscript{2}, 
Ye Zhang\textsuperscript{1}, 
Ziru Yu\textsuperscript{1}, 
Yulan Guo\textsuperscript{1}\thanks{Corresponding author: Yulan Guo}
 \\ \textsuperscript{1}The Shenzhen Campus of Sun Yat-Sen University, Sun Yat-Sen University\\ \textsuperscript{2}Aviation University of Air Force \\ {\tt\small \{liuhd36@mail2.sysu, guoyulan@sysu\}.edu.cn}
}

\begin{document}
\thispagestyle{empty}
\pagestyle{empty}
\maketitle
\thispagestyle{empty}
\pagestyle{empty}
\begin{abstract}
{Global effective receptive field plays a crucial role for image style transfer (ST) to obtain high-quality stylized results. However, existing ST backbones (e.g., CNNs and Transformers) suffer huge computational complexity to achieve global receptive fields.
} 
Recently, State Space Model (SSM), especially the improved variant Mamba, has shown great potential for long-range dependency modeling with linear complexity, which offers an approach to resolve the above dilemma. In this paper, we develop a Mamba-based style transfer framework, termed SaMam. Specifically, a mamba encoder is designed to efficiently extract content and style information. In addition, a style-aware mamba decoder is developed to flexibly adapt to various styles. {Moreover, to address the problems of local pixel forgetting, channel redundancy and spatial discontinuity of existing SSMs, we introduce local enhancement and zigzag scan mechanisms.} Qualitative and quantitative results demonstrate that our SaMam outperforms state-of-the-art methods in terms of both accuracy and efficiency. 
\end{abstract}    
\section{Introduction}
\label{sec:intro}

\begin{figure}[t]
		\centering
		\includegraphics[width=1\linewidth]{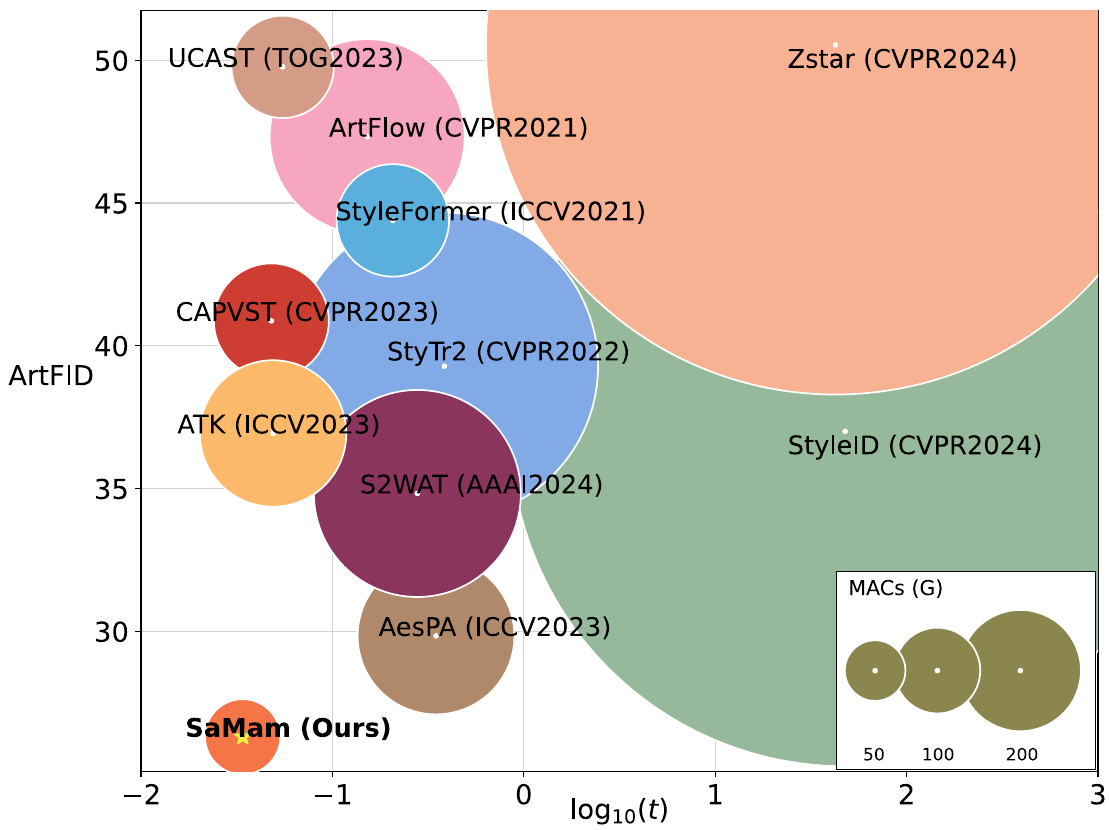}
  \caption{Trade-off between inference time $t$ (ms) and ArtFID~\cite{wright2022artfid} achieved by different methods. The size of a circle represents MACs (G).}
		\label{scatter1}
	\end{figure}

Style transfer (ST) aims at capturing image style to generate artistic images, which has attracted increasing interests since the seminal works~\cite{gatys2015neural,gatys2016image}. With the developments of modern deep learning techniques such as CNNs~\cite{huang2017arbitrary,chandran2021adaptive,hong2023aespa}, transformers~\cite{deng2022stytr2,wu2021styleformer,zheng2024puff}, and diffusion models~\cite{kwondiffusion,deng2024z,chung2024style}, style transformation performance has continued to be improved in the past few years.
We attribute this improvement partly to the increase of receptive fields.
First, a relatively large receptive field allows the model to extract sufficient image patterns from a wider region, enabling it to better capture style patterns~\cite{deng2022stytr2}. Second, with a larger receptive field, the model is able to leverage more pixels in the content image to facilitate the style transformation of the anchor pixel~\cite{guo2024mambair}.

Despite superior performance, previous methods achieve larger receptive fields at the cost of higher computational cost. CNN-based methods stack more convolutional layers to enlarge receptive fields~\cite{simonyan2014very,huang2017arbitrary}, at the cost of high computational overhead. In addition, Transformer-based methods obtains global receptive field at cost of quadratic computational complexity~\cite{deng2022stytr2,guo2024mambair}. For diffusion-based models, as numerous iterations are required, these methods also suffer high computational cost~\cite{chung2024style,deng2024z}.

Recently, a novel State Space Model (SSM) called Mamba~\cite{gu2023mamba} is proposed in the NLP field for long sequence modeling with linear complexity~\cite{fu2022hungry,gu2023mamba,gu2021efficiently,mehta2022long,smith2022simplified}. Mamba introduces an effective solution to balance global receptive field and computational efficiency~\cite{zhu2024vision,liu2024vmambavisualstatespace,guo2024mambair,ma2024u}. Specifically, the discretized space equations in Mamba are formalized into a recursive form and can model long-range dependency when equipped with specially designed structured reparameterization~\cite{gu2023mamba,guo2024mambair}.

In this paper, we propose \textbf{S}tyle-\textbf{a}ware \textbf{Mam}ba (SaMam) ST network, a model to adapt Mamba to balance generation quality and efficiency for ST tasks. First, we design Mamba encoder to efficiently model long-range dependency for image content and style pattern. Second, we propose a Style-aware Mamba decoder. Particularly, we propose a novel Style-aware Selective Scan Structured State Space Sequence Block (S7 block), which efficiently introduces style information to state space updating by predicting weighting parameters in SSM from style embeddings. Furthermore, we design several additional style-aware modules, which incorporates style information to perform feature adaption. Finally, we introduce a zigzag selective scanning method to process image token sequences in a spatially continuous way, which improve semantic continuity. Experiments show our SaMam strikes a better balance between accuracy and efficiency, as illustrated in Fig.~\ref{scatter1}.

In summary, our contributions are three-fold:
\begin{itemize}
    \item We propose SaMam, which balances global effective receptive field with linear computational complexity, making it a good alternative for ST backbones.
    \item We develop a Mamba encoder to extract accurate content features and style patterns. In addition, we design a pluggable style-aware Mamba decoder with flexible adaption to different styles based on learned style embeddings. Moreover, a zigzag scanning method is introduced to obtain superior stylized results.
    \item Extensive experiments shows that our SaMam outperforms other methods in terms of transformation quality and efficiency.
\end{itemize}

\section{Related Work}

\subsection{Neural Style Transfer}

In earlier stage, Gatys \emph{et al.}~\cite{gatys2015neural,gatys2016image} proposed optimization-based methods to obtain stylized images. To achieve faster generation speed, feed-forward methods~\cite{johnson2016perceptual,ulyanov2016texture} are proposed. Specifically, researchers adapt multi-image styles to corresponding network structures~\cite{liu2024pluggable,chen2017stylebank,dumoulin2016learned,zhang2018multi} to enhance the generalization of ST.

More generally, arbitrary ST (AST) attracts increasing interests. Some researchers find that pre-trained CNN models~\cite{szegedy2015going,simonyan2014very} can accurately capture image content and style information. These CNN models are applied to ST as an image feature encoder, which is capable of any image style~\cite{chandran2021adaptive,hong2023aespa,huang2017arbitrary,zhang2022exact,zhu2023all}. Despite the success, CNN-based ST methods typically face challenges in effectively modeling global dependencies. With the development of attention mechanism~\cite{vaswani2017attention}, self-attention is applied in CNN-based ST methods to obtain better stylized results~\cite{zhu2023all,deng2021arbitrary,hong2023aespa}. Furthermore, as transformer have been proven to be a competitive backbone compared to CNN in multiple computer vision tasks~\cite{carion2020end,liu2021swin,dosovitskiy2020image}, researchers apply it in ST~\cite{deng2022stytr2,wu2021styleformer,zheng2024puff,wang2022fine,zhang2024s2wat} to obtain more harmonious stylized results. However, global effective receptive fields comes at the expense of model efficiency. Recently, with developments of diffusion model~\cite{ho2020denoising,nichol2021improved} in generation tasks, researchers utilize it in ST~\cite{kwondiffusion,cheng2023general}. As diffusion-based methods require amount of time to train and synthesize a single image,~\cite{deng2024z,chung2024style} proposes to leverage the generative capability of a pre-trained large-scale diffusion model to improve model efficiency. However, numerous iterations are also required by diffusion-based methods. The dilemma of the trade-off between efficient computation and modeling global dependencies is not essentially resolved.

\subsection{State Space Model}

State Space Model (SSM), as a key component in control theory, is recently introduced to deep learning as a competitive backbone for state space transforming~\cite{gu2021efficiently,smith2022simplified}. Compared to the quadratic complexity of the self-attention mechanism, SSM achieves competitive performance in long sequence modeling with only linear complexity. Structured State Space Sequence model (S4)~\cite{gu2021efficiently} proposes to normalize the parameter matrices into a diagonal structure, which is a seminal work for the deep state space model in modeling the long-range dependency. Furthermore, S5 layer~\cite{smith2022simplified} is proposed based on S4 and introduces MIMO SSM and efficient parallel scan.~\cite{fu2022hungry} designs H3 layer which nearly fills the performance gap between SSM and Transformer attention in natural language modeling.~\cite{mehta2022long} builds the Gated State Space layer on S4 by introducing more gating units to boost the expressivity and accelerate model training. To tackle the limitation of S4 in capturing the contextual information, Gu \emph{et al.}~\cite{gu2023mamba} propose Mamba, which is a novel parameterization method for SSM that integrates an input-dependent selection scan mechanism (\emph{i.e.}, selective scan S4, referred to as S6) and efficient hardware design. Mamba outperforms Transformer on natural language and enjoys linear scaling with input length. Moreover, there are also pioneering works that adopt SSM to vision tasks such as image classification~\cite{zhu2024vision,liu2024vmambavisualstatespace}, image restoration~\cite{guo2024mambair,shi2024vmambair}, biomedical image segmentation~\cite{ma2024u,wang2024large} and others~\cite{nguyen2022s4nd,islam2023efficient,wang2023selective,xie2024fusionmamba}.

\begin{figure*}[t]
		\centering
		\includegraphics[width=0.95\linewidth]{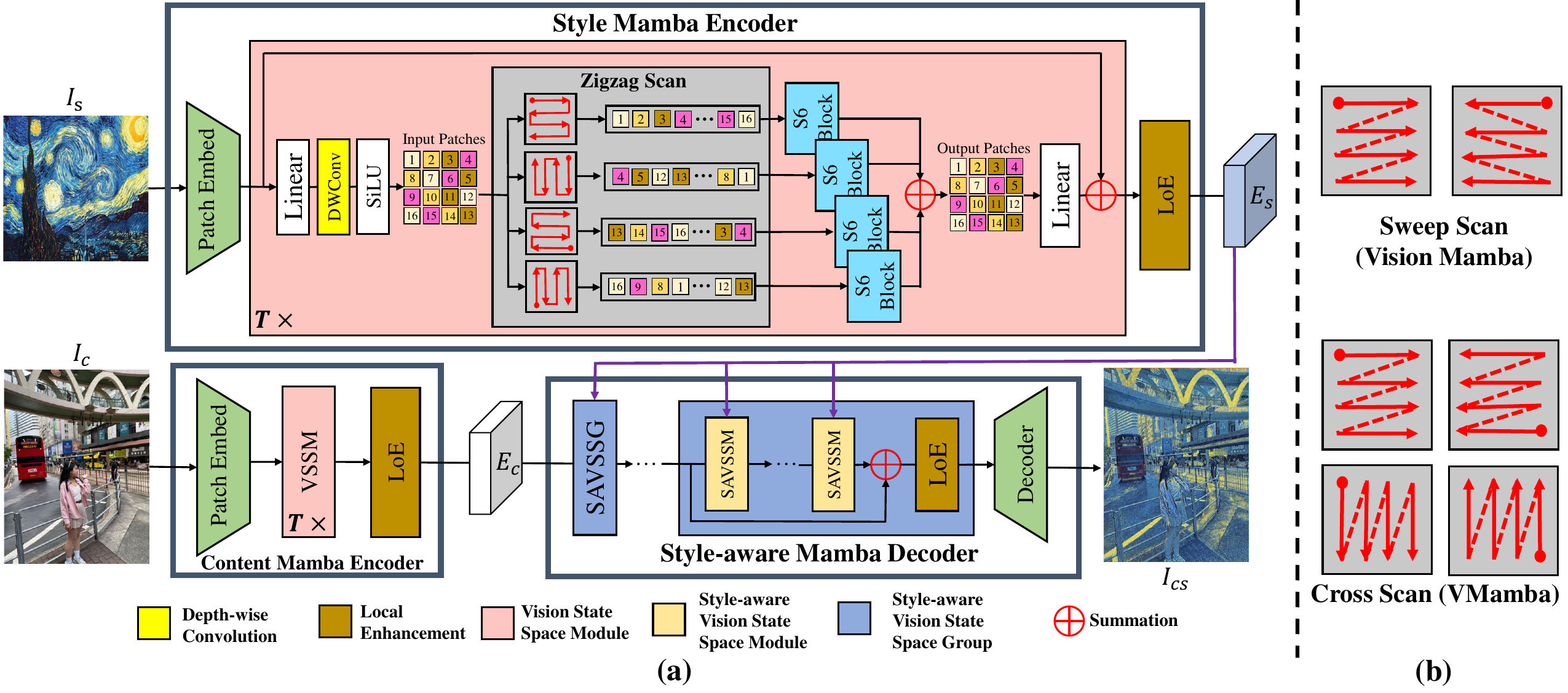}
  \vspace{-12pt}
  \caption{An overview of our SaMam framework (a) and an illustration of the selective scan methods in Vision mamba~\cite{zhu2024vision} and VMamba~\cite{liu2024vmambavisualstatespace} (b).}
		\label{mainnet}
    \vspace{-15pt}
	\end{figure*}
 
\section{Methodology}
\label{Methodology}


\subsection{Preliminaries}

Structured state space sequence models (S4) and Mamba are inspired by the continuous system, which maps a 1-D function or sequence $x(t)\in\mathbb{R}{\rightarrow}y(t)\in\mathbb{R}$ through an implicit latent state $h(t)\in\mathbb{R}^N$. Concretely, continuous-time SSMs can be formulated as linear ordinary differential equations (ODEs) as follows,
\begin{equation}\label{pure_ssm}
\begin{split}
  h'(t) &= \mathbf{A} h(t) + \mathbf{B} x(t),  \\
  y(t)  &= \mathbf{C} h(t) + \mathbf{D} x(t).
\end{split}
\end{equation}
where $\mathbf{A} \in \mathbb{R}^{N\times N}$, $\mathbf{B}\in \mathbb{R}^{N\times 1}$, $\mathbf{C}\in \mathbb{R}^{1\times N}$ and $\mathbf{D} \in \mathbb{R}$ are the weighting parameters.

After that, the discretization process is typically adopted to integrate Eq.~\ref{pure_ssm} into practical deep learning algorithms. The process  introduces a timescale parameter $\mathbf{\Delta}$ to transform the continuous parameters $\mathbf{A}$, $\mathbf{B}$ to discrete parameters $\mathbf{\overline{A}}$, $\mathbf{\overline{B}}$. The commonly used method for transformation is zero-order hold (ZOH), which is defined as follows:
\begin{equation}
\begin{aligned}
\label{eq:zoh}
\mathbf{\overline{A}} &= \exp{(\mathbf{\Delta}\mathbf{A})}, \\
\mathbf{\overline{B}} &= (\mathbf{\Delta} \mathbf{A})^{-1}(\exp{(\mathbf{\Delta} \mathbf{A})} - \mathbf{I}) \cdot \mathbf{\Delta} \mathbf{B}.
\end{aligned}
\end{equation}
After the discretization, the discretized version of Eq.~\ref{pure_ssm} with step size $\mathbf{\Delta}$ can be rewritten in the following RNN form:
\begin{equation}
\begin{aligned}
\label{eq:discrete_lti}
h_t &= \mathbf{\overline{A}}h_{t-1} + \mathbf{\overline{B}}x_{t}, \\
y_t &= \mathbf{C}h_t + \mathbf{D}x_{t}.
\end{aligned}
\end{equation}
Finally, the model computes output through a global convolution. 
\begin{equation}
\begin{aligned}
\label{eq:conv}
\mathbf{\overline{K}} &= (\mathbf{C}\mathbf{\overline{B}}, \mathbf{C}\mathbf{\overline{A}}\mathbf{\overline{B}}, \dots, \mathbf{C}\mathbf{\overline{A}}^{\mathtt{L}-1}\mathbf{\overline{B}}), \\
\mathbf{y} &= \mathbf{\overline{K}} \circledast \mathbf{x} + \mathbf{D} * \mathbf{x},
\end{aligned}
\end{equation}
where $\mathtt{L}$ is the length of the input sequence $\mathbf{x}$, $\overline{\mathbf{K}} \in \mathbb{R}^{\mathtt{L}}$ is a structured convolutional kernel and $\circledast$ denotes convolution operation. As recent advanced SSM, Mamba~\cite{gu2023mamba} proposes S6 to improve 
$\overline{\rm \textbf{B}}$, ${\rm \textbf{C}}$ and $\rm \Delta$ to be input-dependent, thus allowing for a dynamic feature representation.


\subsection{Overall Architecture}

Our SaMam consists of a Style Mamba Encoder, a Content Mamba Encoder and a Style-aware Mamba Decoder, as shown in Fig.~\ref{mainnet}(a). First, the content image $\mathbf{I_c} \in \mathbb{R}^{3 \times 4H \times 4W}$ and style image $\mathbf{I_s} \in \mathbb{R}^{3 \times 4H_s \times 4W_s}$ are first fed to encoders to obtain content feature $\mathbf{E_c} \in \mathbb{R}^{C \times H \times W}$ and style embedding $\mathbf{E_s} \in \mathbb{R}^{C \times H_s \times W_s}$, respectively. Next, $\mathbf{E_s}$ is leveraged as style condition information, which is employed to adapt the decoder parameters. Finally, $\mathbf{E_c}$ is fed to the decoder to obtain stylized image $\mathbf{I_{cs}} \in \mathbb{R}^{3 \times 4H \times 4W}$.

\subsection{Style/Content Mamba Encoder}

The images are first embedded to downscaled image features. Then the image features are fed to Vision State Space Modules (VSSMs) to extract deep features. Moreover, an additional local enhancement (LoE) is introduced at the end of encoders to enhance features extracted from VSSM.

Due to the computational efficiency and long-range modeling ability of SS2D block in VMamba~\cite{liu2024vmambavisualstatespace}, we also follow the \texttt{Linear} $\rightarrow$ \texttt{DWConv} $\rightarrow$ \texttt{SS2D} $\rightarrow$ \texttt{Linear} flow.

\begin{figure*}[t]
		\centering
		\includegraphics[width=0.95\linewidth]{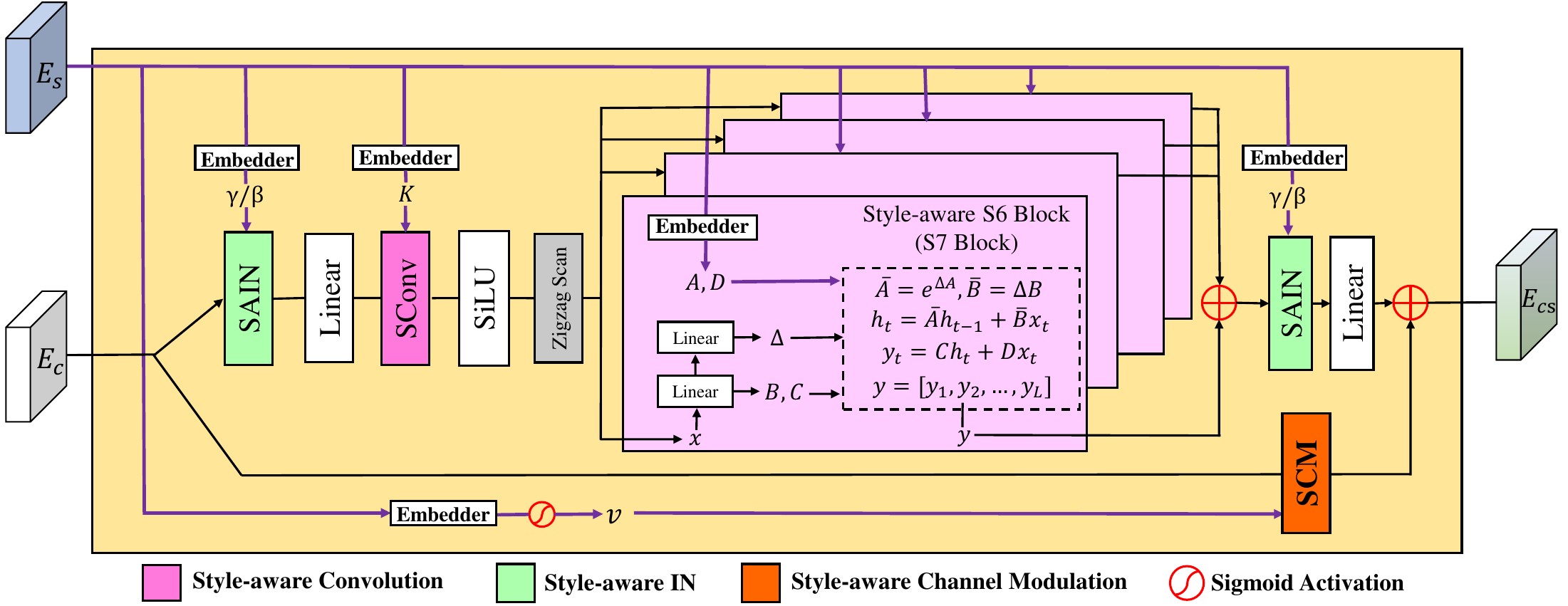}
  \vspace{-10pt}
  \caption{The detailed architecture of Style-aware Vision State Space Module (SAVSSM).}
		\label{SAVSSM}
    \vspace{-15pt}
	\end{figure*}

\noindent\textbf{Zigzag Scan:} Prior researchs~\cite{zhu2024vision,liu2024vmambavisualstatespace} have demonstrated the efficacy of using multiple scanning orders to improve performance (\emph{e.g.}, row-wise and column-wise scans in multiple directions, as shown in Fig.~\ref{mainnet}(b)). Previous scanning order can only cover one type of 2D direction (\emph{e.g.}, left to right), which causes spatial discontinuity when moving to a new row or column~\cite{yang2024plainmamba,zhu2024rethinking}. Moreover, as the parameter $\mathbf{\overline{A}}$ in Eq.~\ref{eq:conv} serves as a decaying term, the spatial discontinuity causes abrupt changes in degrees of decaying of adjacent tokens, compounding the semantic discontinuity and resulting in unnatural stylized textures. Inspired by~\cite{yang2024plainmamba,hu2024zigma}, which proposes Continuous 2D Scanning for semantic continuity, we implement a Zigzag Scan (as shown in Fig.~\ref{mainnet}(a)). The proposed method starts with 4 vertices, with the first clockwise column (or row) as the starting scan-line, aiming at preserving spatial and semantic continuity and generate harmonious stylized results.

\noindent\textbf{Local Enhancement:} As mentioned in~\cite{guo2024mambair}, since SSMs process flattened feature maps as 1D token sequences, the number of adjacent pixels in the sequence is greatly influenced by the flattening strategy. The over-distance in a 1D token sequence between spatially close pixels can lead to local pixel forgetting (\emph{e.g.}, the patch at row $i$ and column $j$ is no longer adjacent to the patch at row $(i + 1)$ and column $j$ in the row-major scan). Moreover, SSMs lead to notable channel redundancy due a larger number of hidden states to memorize very long-range dependencies. To avoid these problems, we add a Local Enhancement (LoE) at the end of VSSM. Specifically, the LoE consists of a convolution layer to compensate for local features and a channel attention layer to facilitate the expressive power of different channels.

\subsection{Style-aware Mamba Decoder}

In decoder, the content feature $\mathbf{E_c}$ and style embedding $\mathbf{E_s}$ are first fed to Style-aware Vision State Space Groups (SAVSSGs) to obtain stylized features $\mathbf{E_{cs}} \in \mathbb{R}^{C \times H \times W}$, and each SAVSSG contains several Style-aware Vision State Space Modules (SAVSSMs). Besides, a LoE is implemented at the end of each SAVSSG to refine features extracted from SAVSSMs. Finally, a lightweight decoder which is similar to~\cite{chen2017stylebank,wang2023microast}, is introduced to generate stylized image $\mathbf{I_{cs}}$.

\subsubsection{Style-aware Vision State Space Module}

The original Mamba Module is designed for the 1-D sequence, which is not suitable for ST tasks requiring spatial-aware understanding. To this end, we introduce the SAVSSM, which incorporates the multi-directional sequence modeling for the vision tasks. Moreover, to achieve flexible style-aware adaption, we propose a style-pluggable mechanism (as shown in Fig.~\ref{SAVSSM}). Specifically, operations of SAVSSM are presented in Algorithm~\ref{alg:SAVSSM}. The style embedding $\mathbf{E_s}$ serves as condition information, which expands to parameters of style-aware structures. Then content feature $\mathbf{E_c}$ is first normalized by Style-aware Instance Norm (SAIN), and linearly projected it to $\mathbf{E'_c}$. $\mathbf{E'_c}$ is next processed by Style-aware Convolution (SConv). Furthermore, we process $\mathbf{E'_c}$ from 4 directions. For each Style-aware S6 Block (S7 Block), We linearly project token sequence $\mathbf{x}$ to the $\mathbf{B}, \mathbf{C}, \mathbf{\Delta}$, respectively. Then $\mathbf{\Delta}$ is used to discrete $\mathbf{B}$ and $\mathbf{A}$ to obtain $\overline{\mathbf{A}}$ and $\overline{\mathbf{B}}$. Then we compute $\mathbf{y}$ through SSM. After that, the outputs are added together and normalized to get the output token sequence $\mathbf{E'_{cs}}$. We linearly project $\mathbf{E'_{cs}}$ and sum it with residual to get stylized feature $\mathbf{E_{cs}}$. Moreover, a Style-aware Channel Modulation (SCM) is implemented in the residual branch.

\begin{algorithm}[t]
\caption{\name{} Process}
\small
\label{alg:SAVSSM}
\begin{algorithmic}[1]
\REQUIRE{
    content feature $\mathbf{E_c}$: \textcolor{shapecolor}{$(\mathtt{C}, \mathtt{H}, \mathtt{W})$},\\
    style embedding $\mathbf{E_s}$: \textcolor{shapecolor}{$(\mathtt{C}, \mathtt{H_s}, \mathtt{W_s})$}
}
\ENSURE{stylized feature $\mathbf{E_{cs}}$: \textcolor{shapecolor}{$(\mathtt{C}, \mathtt{H}, \mathtt{W})$}}

\STATE \textcolor{gray}{\text{/* pre-proces content feature $\mathbf{E_c}$ */}}

\STATE $\mathbf{E'_c}$: \textcolor{shapecolor}{$(\mathtt{C}, \mathtt{H}, \mathtt{W})$} $\leftarrow$ $\mathbf{SAIN}(\mathbf{E_c}, \mathbf{E_s})$

\STATE $\mathbf{E'_c}$ : \textcolor{shapecolor}{$(\mathtt{E}, \mathtt{H}, \mathtt{W})$} $\leftarrow$ $\mathbf{Linear}(\mathbf{E'_c})$

\STATE $\mathbf{E'_c}$ : \textcolor{shapecolor}{$(\mathtt{E}, \mathtt{H}, \mathtt{W})$} $\leftarrow$ $\mathbf{SiLU}(\mathbf{SConv}(\mathbf{E'_c}, \mathbf{E_s}))$

\STATE \textcolor{gray}{\text{/* process with four S7 Blocks, sequence length $\mathtt{L}=\mathtt{H*W}$ */}}
\FOR{$p$ in \{$path1$, $path2$, $path3$, $path4$\}}
    \STATE $\mathbf{x}_p$: \textcolor{shapecolor}{$(\mathtt{L}, \mathtt{E})$} $\leftarrow$ $p(\mathbf{E'_c})$
    
    \STATE $\mathbf{B}_p$: \textcolor{shapecolor}{$(\mathtt{L}, \mathtt{N})$} $\leftarrow$ $\mathbf{Linear}^{\mathbf{B}}_p(\mathbf{x}_p)$
    
    \STATE $\mathbf{C}_p$: \textcolor{shapecolor}{$(\mathtt{L}, \mathtt{N})$} $\leftarrow$ $\mathbf{Linear}^{\mathbf{C}}_p(\mathbf{x}_p)$
    
    \STATE \textcolor{gray}{\text{/* softplus ensures positive $\mathbf{\Delta}_p$ */}}
    
    \STATE $\mathbf{\Delta}_p$: \textcolor{shapecolor}{$(\mathtt{L}, \mathtt{E})$} $\leftarrow$ $\log(1 + \exp(\mathbf{Linear}^{\mathbf{\Delta}}_p(\mathbf{x}_p) + \mathbf{Parameter}^{\mathbf{\Delta}}_p))$

    \STATE \textcolor{gray}{\text{/* style-aware parameters */}}
    
    \STATE $\mathbf{A}_p$: \textcolor{shapecolor}{$(\mathtt{N}, \mathtt{E})$} $\leftarrow$ \textcolor{blue}{$\mathbf{Embedder}^{\mathbf{A}}_p(\mathbf{E_s})$}

    \STATE $\mathbf{D}_p$: \textcolor{shapecolor}{$(\mathtt{E},)$} $\leftarrow$ \textcolor{blue}{$\mathbf{Embedder}^{\mathbf{D}}_p(\mathbf{E_s})$}

    \STATE \textcolor{gray}{\text{/* discretization process */}}

    \STATE $\overline{\mathbf{A}}_p$: \textcolor{shapecolor}{$(\mathtt{L}, \mathtt{N}, \mathtt{E})$} $\leftarrow$ $\exp(\mathbf{\Delta}_p \bigotimes \mathbf{A}_p)$ 
    
    \STATE $\overline{\mathbf{B}}_p$ : \textcolor{shapecolor}{$(\mathtt{L}, \mathtt{N}, \mathtt{E})$} $\leftarrow$ $\mathbf{\Delta}_p \bigotimes \mathbf{B}_p$
    
    \STATE $\mathbf{y}_p$ : \textcolor{shapecolor}{$(\mathtt{L}, \mathtt{E})$} $\leftarrow$ $\mathbf{SSM}(\overline{\mathbf{A}}_p, \overline{\mathbf{B}}_p, \mathbf{C}_p, \mathbf{D}_p)(\mathbf{x}_p)$

    \STATE $\mathbf{y}_p$ : \textcolor{shapecolor}{$(\mathtt{E}, \mathtt{H}, \mathtt{W})$} $\leftarrow$ $\mathbf{Merge}(\mathbf{y}_p)$
    
\ENDFOR

\STATE $\mathbf{E'_{cs}}$ : \textcolor{shapecolor}{$(\mathtt{E}, \mathtt{H}, \mathtt{W})$} $\leftarrow$ $\mathbf{SAIN}(\mathbf{y}_{path1}+\mathbf{y}_{path2}+\mathbf{y}_{path3}+\mathbf{y}_{path4},\mathbf{E_s})$

\STATE $\mathbf{E_{cs}}$ : \textcolor{shapecolor}{$(\mathtt{C}, \mathtt{H}, \mathtt{W})$} $\leftarrow$ $\mathbf{Linear}(\mathbf{E'_{cs}}) + \mathbf{SCM}(\mathbf{E_c}, \mathbf{E_s})$

Return: $\mathbf{E_{cs}}$ 
\end{algorithmic}
\end{algorithm}

\noindent\textbf{Style-aware S6 Block (S7 Block):} Different from $\mathbf{A}$ and $\mathbf{D}$ from a certain concrete embedding space in standard S6 block, we introduce a dynamical weights generation scheme. Specifically, we predict $\mathbf{A}$ and $\mathbf{D}$ from style-embedding $\mathbf{E_s}$:
\begin{equation}
\mathbf{A}, \mathbf{D} = \mathbf{Embedder}(\mathbf{E_s}),
\end{equation}
where $\mathbf{A} \in \mathbb{R}^{E\times N}$, $\mathbf{D} \in \mathbb{R}^{E \times 1}$. $E$ and $N$ represent expanded dimension size and SSM dimension, respectively. We design the S7 block based on 2 aspects. \textbf{(1) Style Selectivity:} Standard S6 block updates hidden state based on content only. However, the hidden state should be affected by both content and style. Furthermore, concrete embedding $\mathbf{A}$ in S6 block could also acquire selectivity~\cite{gu2023mamba} by Eq.~\ref{eq:zoh}. To introduce style information in hidden state updating, we utilize the selectivity of $\mathbf{A}$ by predicting it from style embedding space, instead of concrete embedding. \textbf{(2) Efficiency:} As shown in Eq.~\ref{eq:conv}, weighting parameters $\mathbf{A}$ and $\mathbf{D}$ expand to convolution kernel and channel-wise scale factor, respectively, which is similar to~\cite{wang2021unsupervised,chandran2021adaptive,he2020interactive}. The dynamical global convolution kernel maintains efficient computation by parallel operations while adapting to various styles.

\noindent\textbf{{Additional Style-aware Modules:}} To achieve better visual quality, we also implement several additional style-aware structures to fuse content and style information.

\textbf{(1) SConv:} Inspired by AdaConv~\cite{chandran2021adaptive} that proposes a style-aware depthwise convolution to better preserve local geometric structures of style images, we replace DWConv by SConv. Specifically, style embedding $\mathbf{E_s}$ is passed to an embedder to generate the convolution kernels $K$ in SConv. Note that, $K \in {\mathbb{R}}^{C\times{1}\times{k_w}\times{k_h}}$. Then the predicted convolution kernels $K$ perform depthwise convolution operation on content image feature $\mathbf{E'_c}$.

\textbf{(2) SCM:} Inspired by CResMD~\cite{he2020interactive} that uses controlling variables to rescale different channels to handle multiple image degradations, our SCM learns to generate modulation coefficients based on the style embedding $\mathbf{E_s}$ to perform channel-wise feature adaption. Specifically, $\mathbf{E_s}$ is passed to the embedder and sigmoid activation layer to generate channel-wise modulation coefficients $v \in {\mathbb{R}}^{C}$. Then, $v$ is used to rescale different channel components in $\mathbf{E_c}$.

\textbf{(3) SAIN:} In addition to local geometric structures, the global properties are also critical to the final results. Following the widespread usage of adaptive normalization in ST~\cite{huang2017arbitrary,jing2020dynamic,liu2024pluggable}, visual reasoning~\cite{perez2018film} and image generation~\cite{dhariwal2021diffusion,peebles2023scalable}, we explore replacing standard norm with adaptive norm to transfer global properties from style images. Compared with channel-wise modulation of layer norm in VSS Block~\cite{liu2024vmambavisualstatespace}, the feature-wise modulation of instance norm is more promising in ST field~\cite{huang2017arbitrary,dumoulin2016learned}. So a style-aware instance norm (SAIN) is proposed. Specifically, style embedding $\mathbf{E_s}$ is fed to an embedder to predict the mean $\gamma$ and variance $\beta$ of the style. Moreover, prior work on ResNets has found that initializing each residual block as the identity function is beneficial. For example, \cite{goyal2017accurate} finds that zero-initializing the final batch norm scale factor in each block accelerates large-scale training in the supervised learning setting. DiT~\cite{peebles2023scalable} uses a similar initialization strategy, zero-initializing the layer norm in each block. Inspired by previous explorations, we initialize embedders of SAIN and SCM to output zero-vector. This initializes the SAVSSM as the identity function.

\begin{figure}[t]
		\centering
		\includegraphics[width=1\linewidth]{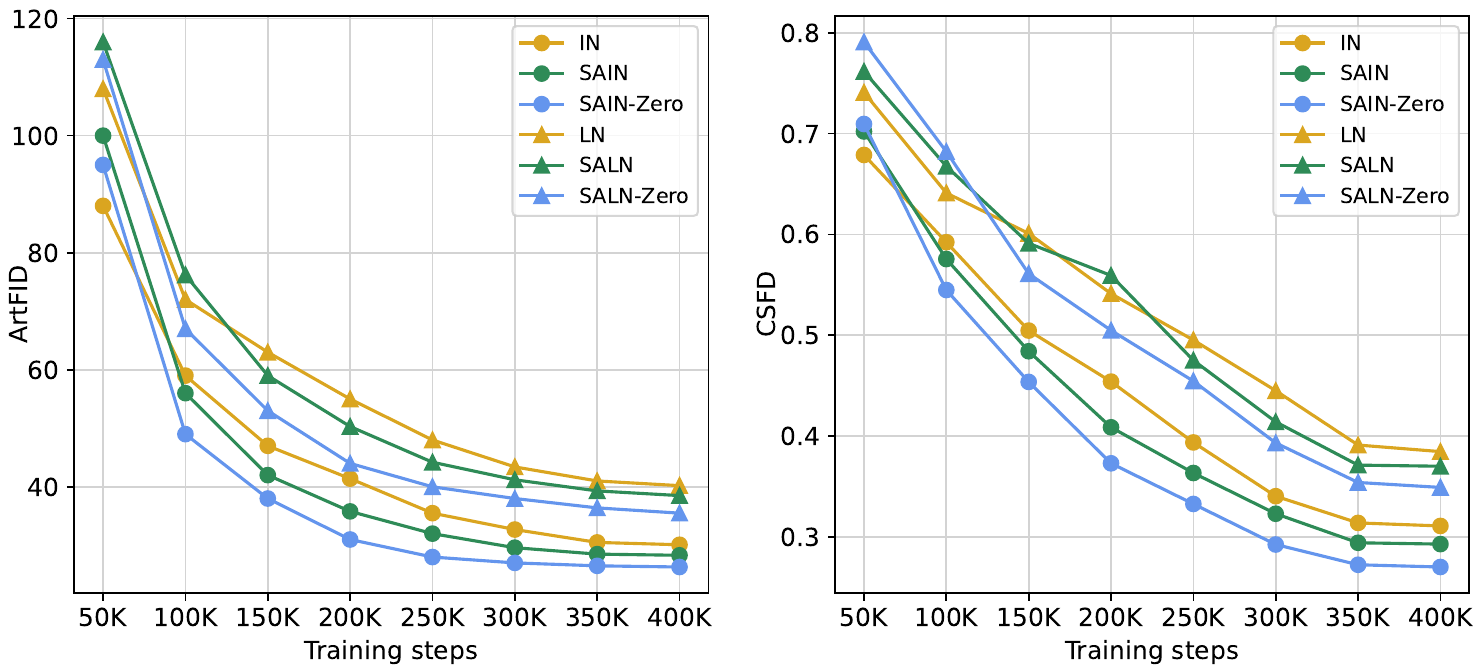}
  \vspace{-20pt}
  \caption{Comparison of different norm strategies.}
		\label{norm_comparison}
    \vspace{-13pt}
	\end{figure}

\begin{figure*}[t]
		\centering
		\includegraphics[width=1\linewidth]{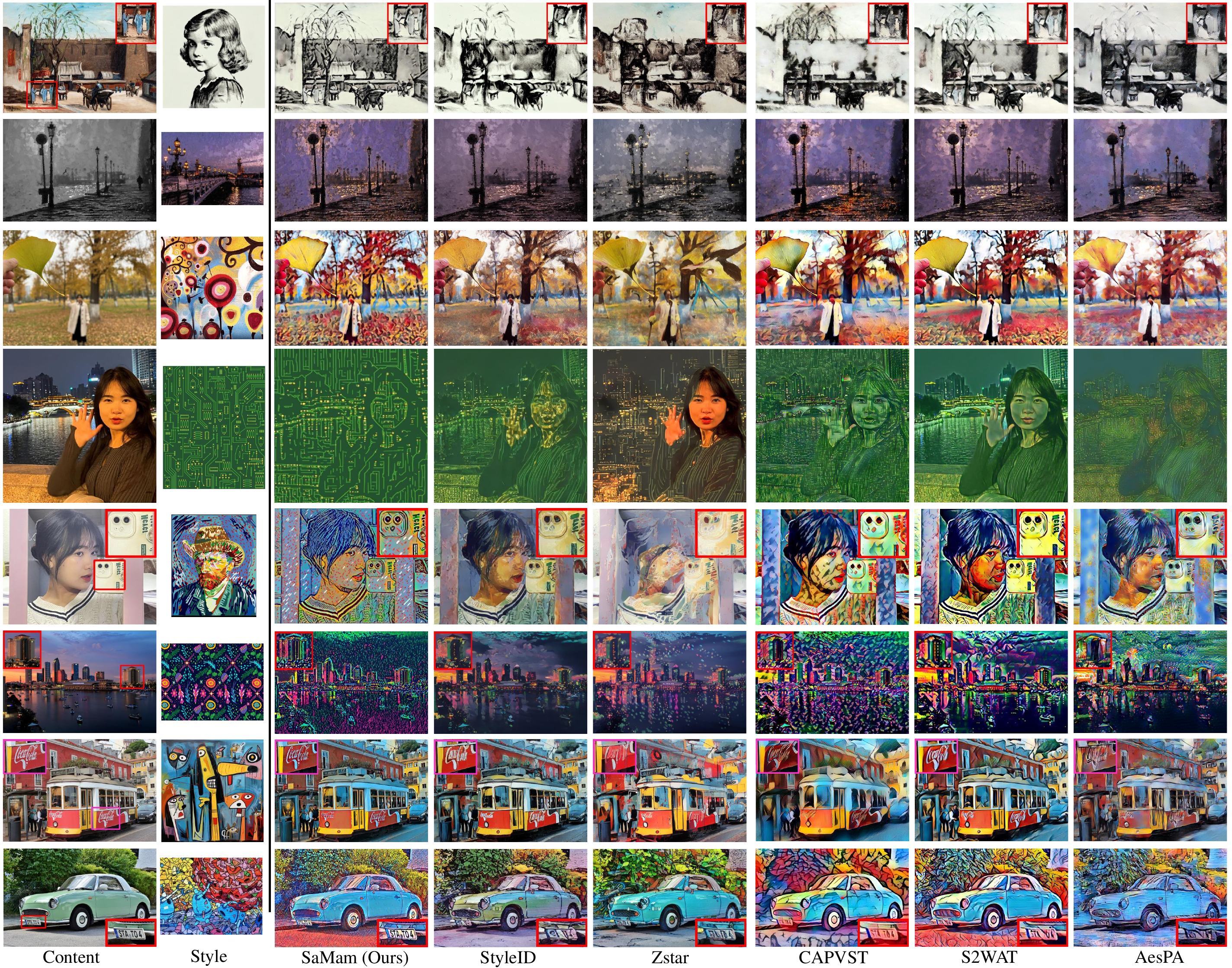}
  \vspace{-20pt}
  \caption{Qualitative comparison with previous state-of-the-art methods.}
		\label{comparison_visual}
	\end{figure*}

\begin{table*}[t]
\vspace{-10pt}
		\caption{Quantitative comparison of the ST methods. The \textbf{best} and \underline{second best} results are highlighted, respectively. Run time and MACs are evaluated on $512\times512$ output resolution with a single NVIDIA RTX 3090 GPU.}
  \vspace{-20pt}
		\begin{center}
                \resizebox{\linewidth}{!}{
				\begin{tabular}{|l|cccc|cccc|cc|cccc|c|}
					\hline 
                        \multirow{2}{*}{Metrics}
					& \multicolumn{4}{c|}{\multirow{1}{*}{CNN based}} 
					& \multicolumn{4}{c|}{\multirow{1}{*}{Transformer based}} 
     & \multicolumn{2}{c|}{\multirow{1}{*}{Reversible-NN based}} 
     & \multicolumn{4}{c|}{\multirow{1}{*}{Diffusion based}}   & \multicolumn{1}{c|}{\multirow{1}{*}{Mamba-based}} 
					\tabularnewline
					
					& AesPA & EFDM  & ATK  & UCAST
					& StyTr2 & S2WAT  & StyleFormer & STTR
                        & ArtFlow & CAPVST
					& DiffuseIT & ZStar & StyleID & VCT
					& SaMam (Ours)
					\tabularnewline
					\hline
					
					LPIPS $\downarrow$
					& \underline{0.4050} & 0.5252 & 0.4320 & 0.5786
					& 0.4992 & 0.4256 & 0.4898 & 0.5755
                        & 0.5671 & 0.4955
					& 0.6954 & 0.5674 & 0.4803 & 0.5429
					& \textbf{0.3884}
					\tabularnewline
     
                        FID $\downarrow$
					& \underline{20.236} & 34.834 & 24.788 & 30.527
					& 25.204 & 23.430  & 28.798 & 30.302
                        & 29.199 & 26.330
					& 37.172 & 31.240 & 24.488 & 37.485
					& \textbf{17.946}
					\tabularnewline

                        ArtFID $\downarrow$
					& \underline{29.837} & 54.655 & 36.928 & 49.768
					& 39.285 & 34.827  & 44.392 & 49.316
                        & 47.325 & 40.871
					& 64.702 & 50.534 & 37.730 & 59.379
					& \textbf{26.305}
					\tabularnewline

                        CFSD $\downarrow$
					& 0.3291 & 0.3391 & 0.3683 & 0.3566
					& 0.4226 & 0.3736 & 0.3976 & 0.3517
                        & 0.3125 & \underline{0.2912}
					& 0.7294 & 0.4277 & 0.3132 & 0.4671
					& \textbf{0.2703}
					\tabularnewline

     \hline
     
					MACs (G) $\downarrow$
					& 334.3 & \textbf{63.3}  & 291.4 & 142.4
					& 1283.5 & 582.6  & 172.1 & 110.4
                        & 517.0 & 179.9
					& - & 6639.0 & 6094.6 & -
					& \underline{77.1}
					\tabularnewline

                        Time (s) $\downarrow$
					& 0.348 & \textbf{0.027} & 0.049 & 0.055
					& 0.385 & 0.278  & 0.207 & 0.181
                        & 0.152 & 0.048
					& 705.214 & 42.439 & 47.746 & 680.972
					& \underline{0.034}
					\tabularnewline
     
                        Params (M) $\downarrow$
                        & 24.20 & 7.01 & 11.18 & 10.52
					& 35.39 & 64.96 & 19.90 & 45.64
                        & \underline{6.46} & \textbf{4.09}
					& 559.00 &  1066.24 & 1066.24 & 1066.24
					& 18.50
					\tabularnewline

					\hline	
			\end{tabular}
   }
		\end{center}
  \label{comparisonwsotas}
  \vspace{-20pt}
	\end{table*}

\subsection{Loss Function}

The overall loss function consists of a content term, a style term, and identity terms, which is defined as follows:
\begin{align}
\vspace{-10pt}
\label{loss_fun}
    \mathcal{L}=\mathcal{L}_{c} + \lambda_{s}\mathcal{L}_{s}+\lambda_{id1}\mathcal{L}_{id1}+\lambda_{id2}\mathcal{L}_{id2}, 
    \vspace{-10pt}
\end{align}
where $\lambda_{s}$, $\lambda_{id1}$ and $\lambda_{id2}$ are set to $10$, $1$ and $50$, respectively.

\noindent\textbf{Content loss and style loss:} Similar to previous works \cite{deng2022stytr2,deng2021arbitrary}, we define content and style loss as follows:
\begin{align}
\mathcal{L}_{c}=\sum_{l \in \{l_c\}}||\phi^l(\mathbf{I_{cs}})-\phi^l(\mathbf{I_c})||_2,
\end{align}
\begin{align}
\mathcal{L}_{s}=\sum_{l \in \{l_s\}}(||\mu(\phi^l(\mathbf{I_{cs}}))-\mu(\phi^l(\mathbf{I_s}))||_2 + \notag \\
||\sigma(\phi^l(\mathbf{I_{cs}}))-\sigma(\phi^l(\mathbf{I_s}))||_2),
\end{align}
where $\phi^l$ refers to features extracted from the $l$-th layer in a pre-trained VGG-19~\cite{simonyan2014very}.  $\mu(\cdot)$ and $\sigma(\cdot)$ denote the mean and variance of extracted features, respectively.

\noindent\textbf{Identity loss:} In order to learn more accurate content and style information, we adopt identity loss~\cite{deng2022stytr2,hong2023aespa}:
\begin{align}
\vspace{-10pt}
\mathcal{L}_{id1}=&||\mathbf{I_{cc}}-\mathbf{I_c}||_2 + ||\mathbf{I_{ss}}-\mathbf{I_s}||_2, \notag \\
\mathcal{L}_{id2}=&\sum_{l \in \{l_{id}\}}(\phi^l(\mathbf{I_{cc}})-\phi^l(\mathbf{I_{c}})||_2 + ||\phi^l(\mathbf{I_{ss}})-\phi^l(\mathbf{I_{s}})||_2),
\vspace{-10pt}
\end{align}
where $\mathbf{I_{cc}}$ (or $\mathbf{I_{ss}}$) refers to the output image synthesized from two with the same content (or style) images.

\section{Experiments}

\subsection{Experimental Setup}

\noindent\textbf{Implementation Details:} We use MS-COCO~\cite{lin2014microsoft} as content dataset and select style images from
WikiArt~\cite{phillips2011wiki}. In Algorithm~\ref{alg:SAVSSM}, image feature channel number $C$, expanded dimension size $E$ and SSM dimension $N$ are set to 256, 512 and 16, respectively. $C$, $E$ and $N$ in VSSM are set the same as those in Algorithm~\ref{alg:SAVSSM}. During training, content and style images are rescaled to $256\times256$ pixels. 8 content-style image patch pairs are randomly selected as a mini-batch. We adopt the Adam optimizer~\cite{kingma2014adam} to train the whole model for $1M$ iterations. The initial learning rate is set to $1\times10^{-4}$ and decreased to half every $0.25M$ iterations.

\noindent\textbf{Evaluation Metrics:} Following the protocol of StyleID~\cite{chung2024style}, we use ArtFID~\cite{wright2022artfid} and content feature structural distance (CFSD) as metrics. Specifically, ArtFID is equal to $(1+\text{LPIPS}) \times (1+\text{FID})$. As the two metrics strongly coinciding with human judgment, LPIPS measures content fidelity while FID assesses the style similarity. Moreover, CFSD is an additional content fidelity metric to measure the spatial correlation between image patches.

\subsection{Comparison with Prior Arts}

We compare our SaMam to recent state-of-the-art ST methods, including CNN based (AesPA~\cite{hong2023aespa}, EFDM~\cite{zhang2022exact}, ATK~\cite{zhu2023all}, UCAST~\cite{zhang2023unified}), Transformer based (StyTr2~\cite{deng2022stytr2}, S2WAT~\cite{zhang2024s2wat}, StyleFormer~\cite{wu2021styleformer}, STTR~\cite{wang2022fine}), Reversible-NN based (ArtFlow~\cite{an2021artflow},CAPVST~\cite{wen2023cap}) and Diffusion based (DiffuseIT~\cite{kwondiffusion}, ZStar~\cite{deng2024z}, StyleID~\cite{chung2024style}, VCT~\cite{cheng2023general}) methods. We obtain the results of the methods by following their official code with default configurations.

\vspace{-10pt}

\subsubsection{Qualitative Comparison}

We show the visual comparisons in Fig.~\ref{comparison_visual}. It can be observed that our SaMam captures global properties (\emph{e.g.}, textures and colors) from style images (\emph{e.g.}, the $1^{st}-4^{th}$ rows), while it also pays attention on local geometry of style patterns (\emph{e.g.}, speckles in the $5^{th}$ row and $6^{th}$ row). In addition to sufficient style information, our method keeps content structures accurately (\emph{e.g.}, the buildings in the $6^{th}$ row) and produces more clear details and achieves higher perceptual quality (\emph{e.g.}, the texts in the $7^{th}$ row and license plate in the $8^{th}$ row). In contrast, StyleID and Zstar destroy content details severely (\emph{e.g.}, the manga portrait in the $1^{st}$ row and the girl's face in the $5^{th}$ row). Although CAPVST is good at capturing colors of style images, it breaks local geometric structures and content details in stylized results (\emph{e.g.}, the $7^{th}$ row). S2WAT and AesPA are also hard to achieve satisfied results.

\vspace{-10pt}

\subsubsection{Quantitative Comparison}

We resort to some quantitative metrics to better evaluate the proposed method. Note that, the MACs of DiffuseIT and VCT are not reported since they conduct training during the inference time and require considerable computational cost.

\noindent\textbf{(1) Stylization Quality:}  We collect 20 content images and 40 style images to synthesize 800 stylized images for each method and show their average metric scores in Table~\ref{comparisonwsotas}. It can be observed that Diffusion based methods face great challenges to balance the content and style. To better fuse content and style features, previous CNN (\emph{e.g.}, AesPA~\cite{hong2023aespa} and ATK~\cite{zhu2023all}) and Transformer based methods utilize attention mechanism to build long-range dependency to extract structural information~\cite{deng2022stytr2}. However, the mechanism poses great challenges to these methods to extract complete image properties 
(\emph{e.g.}, local geometry and details) and generate satisfied results. In contrast, in addition to long-range dependency by Mamba, we design more style-aware architectures (\emph{e.g.}, SConv), which adapt to various styles more flexibly. Then our SaMam achieves best results on the 4 quality metrics, indicating that it can transfer sufficient style patterns while better preserving the content details.

\noindent\textbf{(2) Efficiency:} As shown in Fig.~\ref{scatter1} and Table~\ref{comparisonwsotas}, our SaMam achieves the notable performance gains with competitive computation quantity and inference time. Diffusion based methods require amount of time for DDIM inversions and sampling costs, or even more time to train on a single style. Transformer and Reversible-NN based methods are also time-consuming. In contrast, in terms of MACs and inference time, our method is second only to those of a CNN based method (\emph{i.e.}, EFDM). This is because our Mamba based method performs a global convolution, which processes each image token in a parallel way. The advanced scheme inherits the high inference efficiency of CNN based methods while maintaining long-range dependency. This further demonstrates the superiority of our method.

\subsection{Model Analysis}

\begin{table}[t!]

\caption{Ablation study on proposed components. \emph{rp.b.} stands for ``replace by" and \emph{r.m.} stands for ``remove".}
\vspace{-10pt}

\centering

\resizebox{\linewidth}{!}{
\begin{tabular}{cl|cccc}
\toprule
& \textbf{Configuration} & ArtFID & FID & LPIPS & CSFD \\
\bottomrule

\arch{a} & Ours & \textbf{26.305} & \textbf{17.946} & \textbf{0.3884} & \textbf{0.2703} \\

\bottomrule

\arch{b} & Zigzag Scan \emph{rp.b.} Cross Scan & 26.808 & 18.293 &  0.3895 & 0.2955\\

\arch{c} & \emph{r.m.} LoE  & 28.607 & 19.257 & 0.4122 & 0.2970\\

\bottomrule

\arch{d} & S7 Block \emph{rp.b.} S6 Block   & 30.476 & 20.905 & 0.3913 & 0.2756 \\

\arch{e} & SConv \emph{rp.b.} DWConv & 31.446 & 21.044 & 0.4265 & 0.2973 \\

\arch{f} & \emph{r.m.} SCM     & 29.341 & 20.066 & 0.3928 & 0.3235 \\

\arch{g} & SAIN-zero \emph{rp.b.} IN & 30.112 & 20.435 & 0.4048 & 0.3110 \\

\bottomrule
\end{tabular}
}
\label{tab_ablation_study}
\end{table}

\begin{figure}[t]
\vspace{-10pt}
		\centering	\includegraphics[width=1\linewidth]{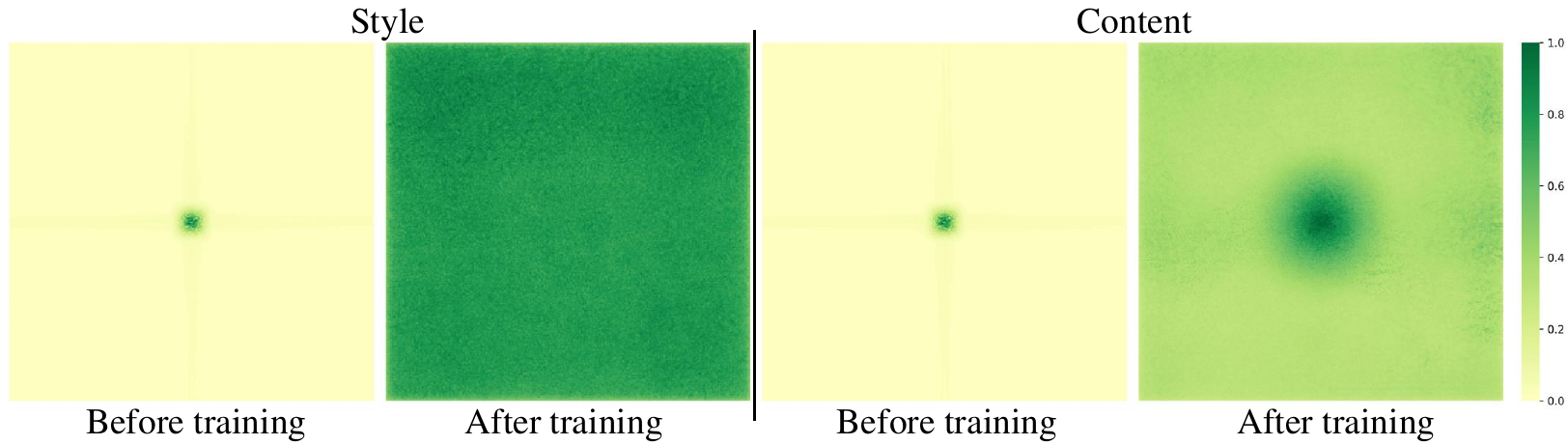}
  \vspace{-13pt}
  \caption{The effective receptive field (ERF) visualization for our SaMam.}
		\label{ERF}
    \vspace{-15pt}
	\end{figure}

\noindent\textbf{Effective Receptive Field (ERF):} We show ERF~\cite{luo2016understanding} in Fig.~\ref{ERF}. A larger ERF is indicated by a more extensively distributed dark area. It can be observed that our SaMam showcase global ERF to capture long-range dependency in terms of style and content after training.

\begin{figure}[t]
		\centering
		\includegraphics[width=1\linewidth]{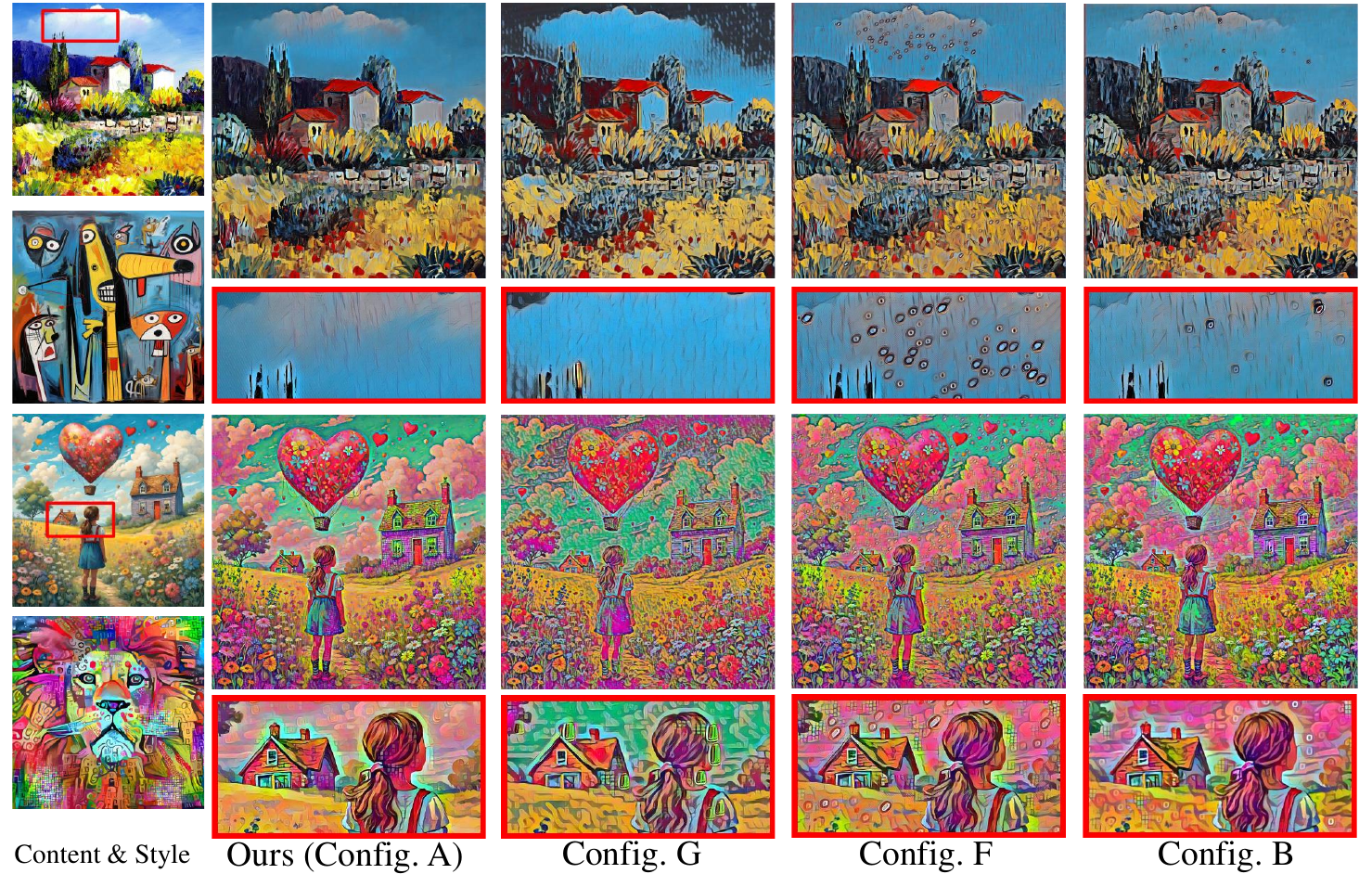}
  \vspace{-20pt}
  \caption{Ablation study on different model configurations.}
    \vspace{-10pt}
		\label{ablation_sadesigns}
	\end{figure}

\noindent\textbf{Zigzag Scan:} To maintain spatial continuity during the scan process, we implement a four-direction zigzag scan method. To demonstrate its effectiveness, we replace zigzag scan by another four-direction scan method (\emph{i.e.}, cross scan) to obtain config.~B. As shown in Table~\ref{tab_ablation_study}, cross scan method jointly reduces ArtFID and CSFD. We further provide visual results in Fig.~\ref{ablation_sadesigns}. It can be observed that zigzag scan method produces clearer background and less artifacts that closed to the content image. This is because that the spatial continuity does not bring abrupt changes to content information, which makes it more difficult to adapt SSM parameters.

\begin{figure}[t]
		\centering
		\includegraphics[width=1\linewidth]{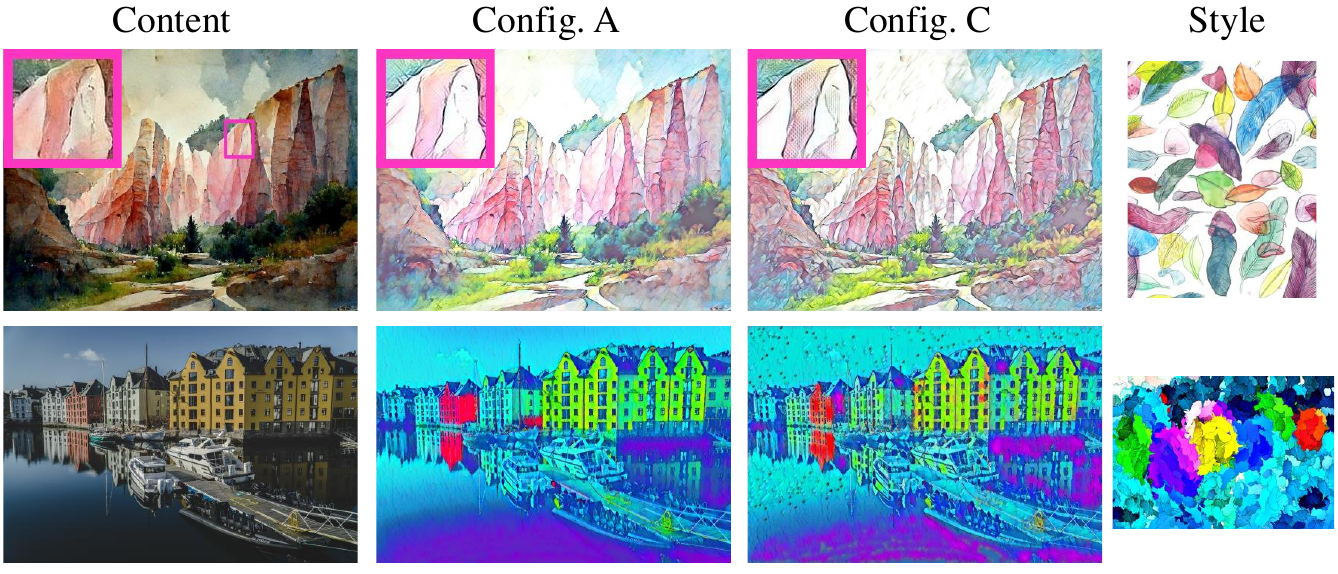}
 \vspace{-20pt}
  \caption{Ablation study on local enhancement. Please zoom in for best view.}
  \vspace{-10pt}
		\label{ablation_LoE}
	\end{figure}

\begin{figure}[t]
		\centering
		\includegraphics[width=1\linewidth]{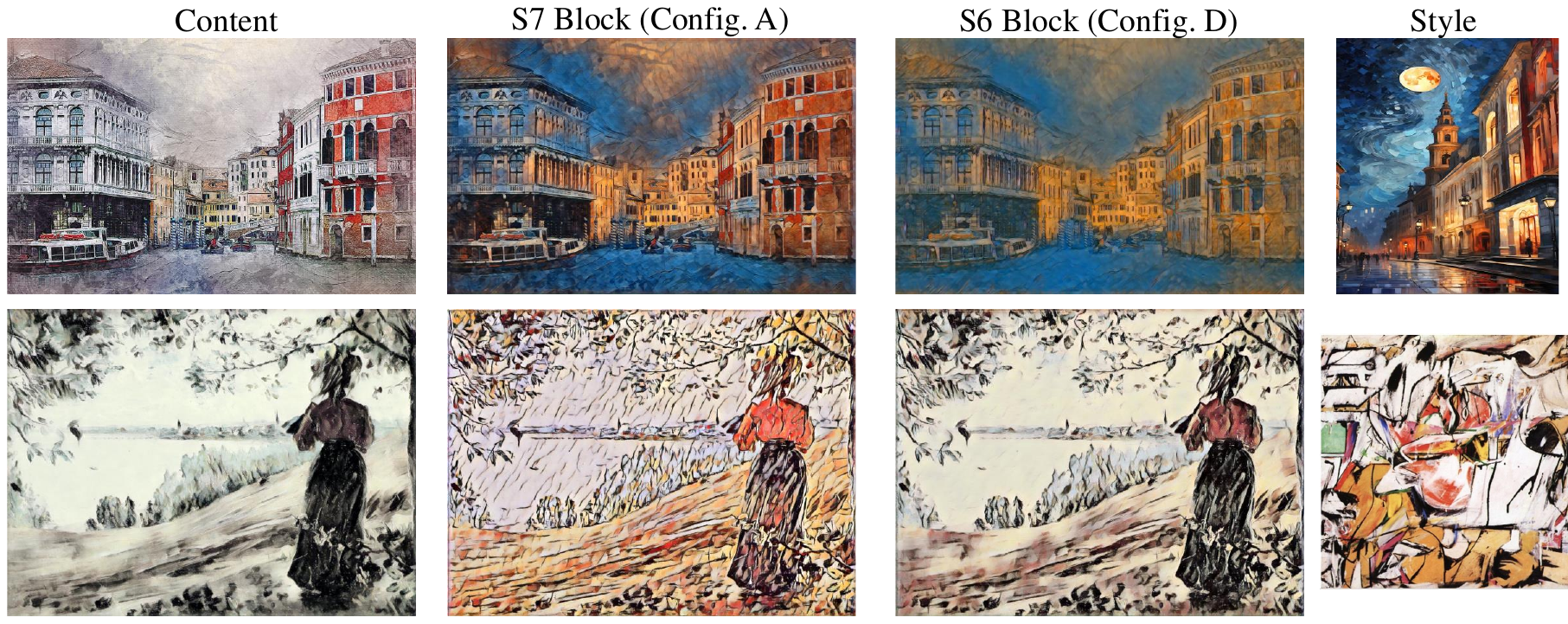}
 \vspace{-20pt}
  \caption{Ablation study on S7 Block.}
  \label{ablation_S7B}
    \vspace{-15pt}
	\end{figure}

\noindent\textbf{Local Enhancement:} We introduce a local enhancement (LoE) module to ease local pixel forgetting. In Fig.~\ref{ablation_LoE}, it can be observed that Config.~C suffers from unnatural noise artifacts which breaks image content smoothness. Quantitative results in Table~\ref{tab_ablation_study} also demonstrate the effectiveness of the LoE.

\noindent\textbf{SAVSSM:} We design a style-aware VSSM to flexibly adapt to different styles. And we further demonstrate the effectiveness of the proposed components.

\textbf{(1) S7 Block:} A novel S7 block is proposed to better capture style properties. We replace S7 block with S6 block to obtain config.~D. It can be observed that our S7 block helps achieve significant higher scores than config.~D. From Fig.~\ref{ablation_S7B}, config.~A reproduces the image color and contrast more closed to the style image. Moreover, it preserves content details and produces sharper edges of higher perceptual quality (\emph{e.g.}, the $1^{st}$ scene in Fig.~\ref{ablation_S7B}). The S7 block inherits global effective receptive field to adapt to various styles.

\begin{figure}[t]
		\centering
		\includegraphics[width=1\linewidth]{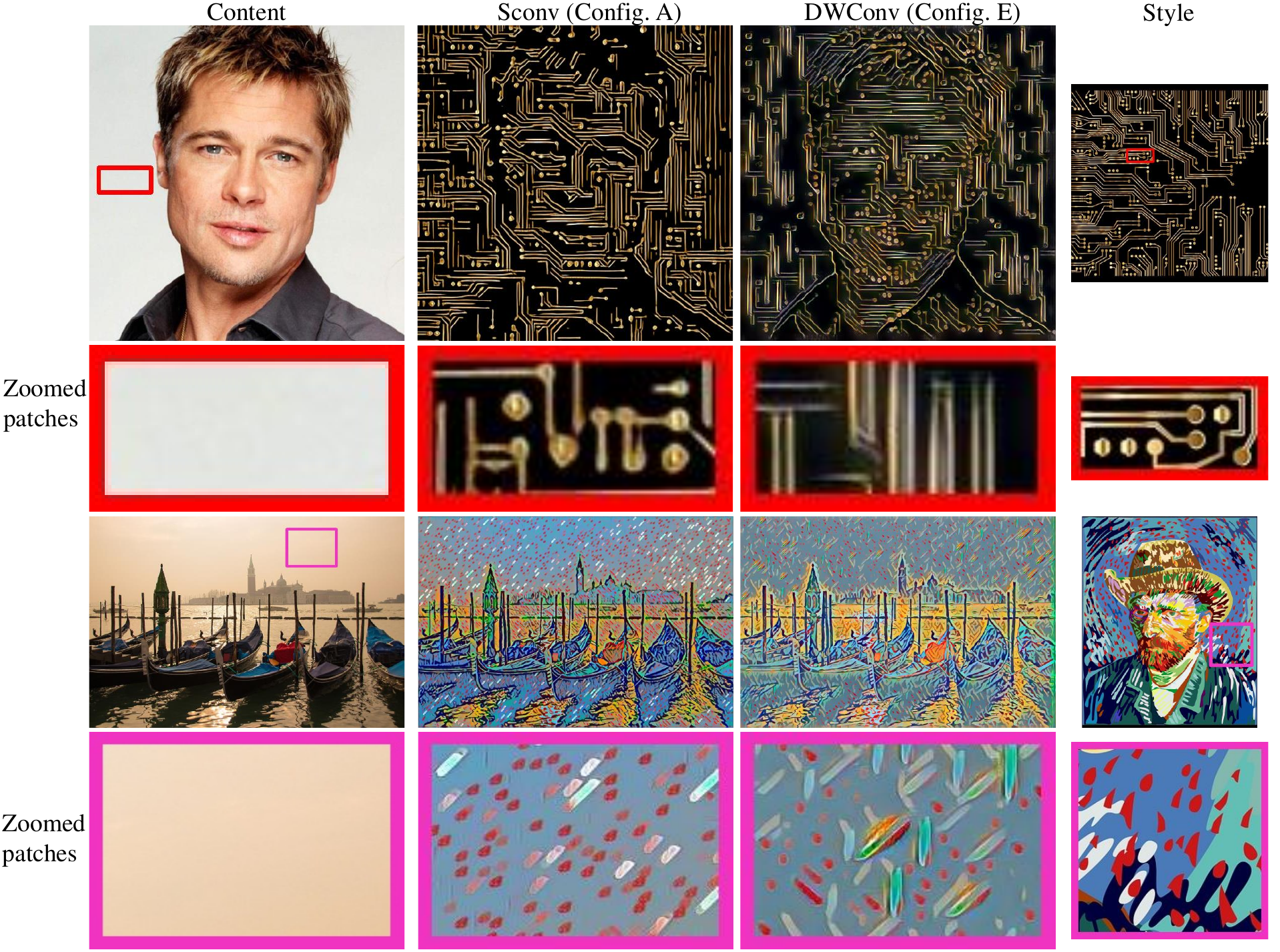}
  \vspace{-16pt}
  \caption{Ablation study on SConv.}
    \label{ablation_sconv}
    \vspace{-15pt}
	\end{figure}

\textbf{(2) SConv:} SConv is proposed to reproduce local geometric structures from style to content. To validate its effectiveness, we replace SConv by common depth-wise convolution (DWConv) layer to obtain config.~E. SConv helps achieve significantly better metrics. For instance, in the first scene of Fig.~\ref{ablation_sconv}, our SConv produces circuit patterns that more closed to the style image.

\textbf{(3) SCM:} As SCM is added to residual branch, our method achieves better visual quality. Moreover, config.~F  achieves significant lower content scores, which indicates that stylized images suffer from more image distortion (\emph{e.g.}, first scene in Fig.~\ref{ablation_sadesigns}).

\textbf{(4) SAIN:} We measure ArtFID and CSFD of various norm strategies during training. Figure.~\ref{norm_comparison} shows the results, which indicates that SAIN with zero initializing outperforms other strategies. Then SAIN-zero is employed in our method to capture global properties from style images. To demonstrate its effectiveness, we replace SAIN with common IN in our SaMam to obtain config.~G, which suffers from significant artifacts on highly chromatic edges (\emph{e.g.}, the second scene in Fig.~\ref{ablation_sadesigns}).

\vspace{0pt}
\section{Conclusion}

In this paper, we explore the power of the recent advanced State Space Model (\emph{i.e.}, Mamba), for arbitrary image style transfer. To this end, we propose a Style-aware Mamba (SaMam) model to strike a trade-off between computational efficiency and global effective receptive field. Specifically, we introduce a Mamba encoder and style-aware Mamba decoder. In addition, we design a style-aware VSSM (SAVSSM) with flexible adaption to various styles based on the style embeddings. Experimental results show that our model achieves state-of-the-art performance for ST task.

\vspace{-5pt}
\section{Acknowledgement}
\vspace{-5pt}

This work was partially supported by the National Natural Science Foundation of China (No. U20A20185, 62372491, 62301601), the Guangdong Basic and Applied Basic Research Foundation (No. 2022B1515020103, 2023B1515120087), the Shenzhen Science and Technology Program (No. RCYX20200714114641140), the Science and Technology Research Projects of the Education Office of Jilin Province (No. JJKH20251951KJ), and the SYSU-Sendhui Joint Lab on Embodied AI.

{
    \small
    \bibliographystyle{ieeenat_fullname}
    \bibliography{main}
}

\end{document}